\documentclass[conference]{IEEEtran}
\IEEEoverridecommandlockouts    % to override locked commands

\usepackage{times}
\usepackage{microtype}
\usepackage{graphicx}
\usepackage{booktabs}
\usepackage{multirow}
\usepackage{amsmath,amssymb}
\usepackage{bm}
\usepackage{siunitx}
\usepackage{hyperref}
\usepackage{cleveref}
\usepackage{xcolor}
\usepackage{placeins}

\newcommand{\R}{\mathbb{R}}

\newcommand{\homo}{\mathbf{A}}
\newcommand{\xbev}{\mathbf{x}}
\newcommand{\pimg}{\mathbf{p}}

\title{\LARGE \bf On-the-Fly Homographies Calibration for Multi-Camera Tracking}

\author{
\IEEEauthorblockN{David Voihanski and Mor Sinai}
\IEEEauthorblockA{\textit{Juganu}\\
Or Yehuda, Israel \\
\{davidv, mors\}@juganu.com}
\and
\IEEEauthorblockN{Ben Zion Bobrovsky}
\IEEEauthorblockA{\textit{Tel Aviv University}\\
Israel \\
bobrov@tauex.tau.ac.il}
}

\IEEEaftertitletext{%
\begin{minipage}{\textwidth}
\scriptsize
\centering
\copyright\ 2026 IEEE. Personal use of this material is permitted.
Permission from IEEE must be obtained for all other uses, in any current
or future media, including reprinting/republishing this material for
advertising or promotional purposes, creating new collective works,
for resale or redistribution to servers or lists, or reuse of any
copyrighted component of this work in other works.
\end{minipage}
\vspace{0.5\baselineskip}
}

\date{}

\begin{document}
\maketitle

\begin{abstract}
Precise multi-camera tracking traditionally relies on rigorous 3D site calibration, yet this requirement is often operationally impossible in large-scale deployments. Privacy regulations frequently prohibit recording video for offline calibration; limited bandwidth precludes synchronizing high-resolution streams from hundreds of cameras; and covering immense physical sites with calibration targets is logistically infeasible.

We present a multi-camera homography calibration system designed to overcome these barriers through "on-the-fly" geometric refinement. Starting from coarse manual homographies, we introduce a centroid-based projection optimization (PO) that continuously aligns the ground-plane geometry using live detection streams. Because PO operates asynchronously on already-transmitted, lightweight metadata, it adds zero computational latency to the real-time tracker. This allows the system to adapt automatically to camera movements or environmental changes without human intervention. This optimized geometry feeds a multi-camera bird's-eye-view (BEV) tracker that fuses detections and unifies trajectories across zones. Crucially, by operating strictly on live anonymous metadata, our solution ensures a privacy-safe, zero-overhead, and resilient tracking pipeline that maintains global consistency in dynamic environments where static, recorded-video calibration is impossible.
\end{abstract}

% ============================================================
\section{Introduction}
\label{sec:intro}
Multi-camera tracking enables global identity consistency across camera networks, improving robustness under occlusion and expanding coverage.
However, many multi-camera systems assume accurate 3D calibration (intrinsics/extrinsics, surveyed landmarks, or specialized procedures). In real-world deployments, relying on such precision is not merely expensive; it is often operationally infeasible~\cite{hartley2003multiple}.

Three primary barriers prevent the adoption of traditional calibration in large-scale facilities. First, \textbf{privacy regulations} (e.g., GDPR) often strictly prohibit the recording and storage of video footage required for offline calibration bundles. Second, \textbf{bandwidth constraints} in large retail or industrial networks preclude the transmission and synchronization of high-resolution video streams from hundreds of cameras to a central server. Third, covering thousands of square meters with physical calibration targets (e.g., checkerboards) is logistically impossible without disrupting site operations.

We target deployments where operators can provide only \emph{manual homographies} from image coordinates to a site map.
Such homographies are fast to obtain but are typically imperfect due to occlusions, imperfect floor plans, and extrapolation errors.
Crucially, static homographies fail when cameras are inevitably nudged or moved during daily operations.

\textbf{Key idea.} We propose a system that requires neither video recording nor physical site access.
By refining initial manual homographies using a projection optimization (PO) procedure that operates on \emph{live detection metadata}, we align multi-camera projections in BEV space "on-the-fly." This approach ensures privacy compliance and allows the system to autonomously heal geometric alignment errors caused by camera movement.

\paragraph{Contributions}
\begin{itemize}
  \item A \textbf{zero-overhead projection optimization} method that refines manual homographies asynchronously. By running exclusively on lightweight, clustered cross-view detection metadata, it continuously adapts to environmental changes without adding computational latency to the real-time pipeline.
  \item A \textbf{global BEV multi-camera tracker} that associates multi-view observations using combined motion and appearance cues before fusing them into a shared geometric state, operating purely on privacy-safe metadata.
  \item A \textbf{trajectory-level cross-camera unification} stage that merges duplicate global tracks using temporal alignment, motion dynamics, and appearance costs.
\end{itemize}

\paragraph{Scope of Tracking Framework}
It is important to note that while we present an end-to-end tracking pipeline, our core scientific contribution is the on-the-fly Projection Optimization (PO) calibration module, not a novel tracking architecture. The custom BEV tracker described herein is provided strictly as an evaluation vehicle to demonstrate the tangible improvements in trajectory alignment resulting from our geometric refinement. Because PO operates purely on the coordinate mapping layer, any multi-camera tracker that relies on spatial gating or trajectory alignment will inherently benefit from the restored geometric fidelity our module provides.

% ============================================================
\section{Related Work}
\label{sec:related}

A well-structured multi-camera tracking system must balance geometric precision with deployment feasibility. Below, we contextualize our approach against the three primary calibration paradigms in the tracking literature, highlighting their operational shortcomings and how our proposed system explicitly addresses them.

\paragraph{Multi-Camera Tracking with 3D Calibration}
Previous work often assumes an accurate 3D camera calibration to triangulate targets or fuse observations in a shared 3D space~\cite{ristani2016performance,tang2019cityflow}. Obtaining such precision in the wild typically requires costly manual setup or specialized procedures, such as surveying architectural landmarks or placing physical checkerboards across the site~\cite{hartley2003multiple}. 
\textbf{Failure Mode:} These systems are inherently brittle in active surveillance environments. If a camera is nudged or moved during daily operations, the rigorous static calibration is immediately invalidated, requiring a costly site visit to repeat the setup.
\textbf{Our Solution:} Our system abandons the need for static, surveyed 3D calibration. By continuously running our Projection Optimization (PO) on live metadata, the system autonomously heals geometric alignment errors on-the-fly, gracefully recovering from camera movements without human intervention.

\paragraph{Planar (Homography) Multi-Camera Tracking}
To bypass full 3D calibration, projecting multi-view observations to a common ground plane (bird's-eye view) using planar homographies is a standard approach for fusing data~\cite{khan2008tracking, fleuret2008multicamera}. 
\textbf{Failure Mode:} These systems typically assume the provided homographies are highly precise. In practice, manual homographies frequently exhibit residual misalignment due to extrapolation errors, lens distortion, and imperfect floor plans~\cite{hartley2003multiple}. This residual spatial scatter is catastrophic for tracking: it causes multi-view observations of the same target to fail spatial distance gating, severely fragmenting trajectories and causing rampant identity switches.
\textbf{Our Solution:} We treat manual homographies merely as a coarse warm start. By dynamically clustering cross-view Re-ID correspondences, our centroid-anchored optimizer continuously pulls the projection matrices into alignment. This drastically reduces spatial scatter and restores the integrity of the tracking system's spatial gating logic.

\paragraph{Weak Calibration and Self-Calibration}
Methods for learning cross-view alignment or self-calibration from data typically attempt to deduce geometry by observing moving targets over time~\cite{dubska2014fully}.
\textbf{Failure Mode:} These approaches generally require strong environmental assumptions (e.g., vehicles moving in perfectly straight lines to find vanishing points), centralized video processing, and long sequences of recorded footage to converge. While effective for traffic surveillance, these assumptions fail completely in retail or pedestrian environments where targets move erratically. Furthermore, in large-scale deployments, strict privacy regulations (e.g., GDPR) and bandwidth constraints prohibit the recording and transmission of the raw video required for these offline calibration algorithms.
\textbf{Our Solution:} Our approach differs by being fully online and privacy-compliant. It relies exclusively on lightweight, anonymized detection metadata to refine the geometry via a stable BEV consistency objective. This eliminates the need for video transmission entirely, making it ideal for bandwidth-constrained, privacy-sensitive networks.

% ============================================================
\section{System Overview}
\label{sec:overview}

At each time step, each camera produces detections (bounding boxes, confidence scores, and appearance embeddings).
A chosen image point (e.g. leg point / bottom-center) is mapped to BEV via a per-camera projection.
Crucially, the edge nodes transmit \emph{only} this lightweight metadata to the central tracker, not video frames. This design respects strict privacy policies regarding video persistence and minimizes bandwidth usage.
We then (i) match per-camera detections to shared BEV tracks, (ii) fuse the matched spatial detections and appearance embeddings to continuously update the global track states, and (iii) periodically merge duplicate BEV tracks via trajectory unification.

% ============================================================
\section{Method}
\label{sec:method}

% ----------------------------
\subsection{Projection from Manual Homographies}
\label{sec:proj_init}
For camera $i \in \{1,\dots,N\}$, we initialize a projective mapping (homography) $\homo_i^{(0)}$ from the image coordinates to a shared BEV plane~\cite{hartley2003multiple}. Given an image point $\pimg \in \R^2$, we compute its projection as $\xbev = \Pi(\pimg; \homo_i) \in \R^2$, where $\Pi$ denotes the homogeneous projection followed by dehomogenization.

In practice, $\mathbf{A}_i^{(0)}$ is obtained from a small set of manual correspondences. While fast to annotate, these suffer from well-known critical limitations: (i) \textbf{down-scaling} reduces point placement accuracy, (ii) \textbf{occlusions} force landmark estimation, (iii) \textbf{imperfect floor plans} mismatch physical reality, (iv) \textbf{floor-contact ambiguity} arises due to perspective differences, (v) \textbf{extrapolation errors} grow significantly outside the annotated region, and (vi) \textbf{lens distortion} cannot be perfectly modeled by simple projective transforms. In a static system, these errors are permanent; in our system, they serve merely as a warm start for online optimization.

\subsection{Matching Point Acquisition}
\label{sec:point_acquisition}

The projection optimization described in Section~\ref{sec:po} relies on clusters of corresponding points $\{\mathcal{C}_c\}$ observed across multiple views. To acquire these correspondences in live deployments where video recording is prohibited or infeasible, we employ an online matching method driven by visual Re-Identification (Re-ID). We run an appearance-based tracker across the camera network to associate detections based on embedding similarity.

Crucially, because projection optimization is highly sensitive to outliers, we prioritize \textbf{precision over recall}. We apply a strictly low matching distance threshold (or high similarity score), rejecting any ambiguous associations. This ensures that we harvest a smaller set of highly robust correspondences rather than a large volume of noisy matches, preventing false associations from corrupting the spatial alignment.

This online approach is critical for \textbf{"on-the-fly" adaptation}. If a camera is accidentally nudged, the Re-ID stream continues to find correspondences between the (now shifted) view and its neighbors. The subsequent optimization step naturally pulls the projection matrix back into alignment without requiring a site visit or manual recalibration, ensuring long-term system resilience.

\paragraph{Spatial Subsampling}
Since the transformation between the image plane and the ground plane is modeled as a homography (a smooth, planar mapping), dense clusters of points within a small image neighborhood provide redundant geometric constraints. To improve computational efficiency and prevent overfitting to local clusters, we apply spatial subsampling on the image plane. We enforce a configurable minimum pixel distance between selected points. This ensures a uniform distribution of constraints across the field of view without unnecessary redundancy.

% ----------------------------
\subsection{Projection Optimization (PO)}
\label{sec:po}

\begin{figure}[htbp]
    \centering
    \includegraphics[width=0.82\linewidth]{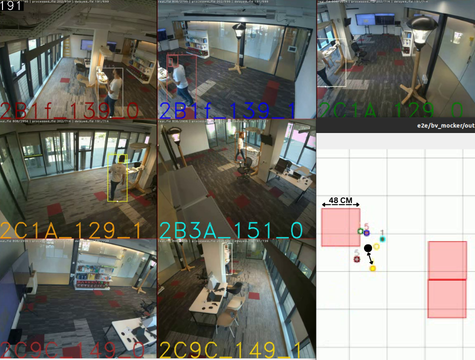}
    \caption{Projection inconsistency before optimization. A single person is observed simultaneously across multiple overlapping camera views (left). When mapped into the shared bird's-eye view (bottom right) using initial manual homographies, the individual camera projections (colored circles) exhibit significant spatial scatter around their computed centroid (black dot). The black arrow illustrates the multi-view alignment error that our projection optimization aims to minimize per camera.}
    \label{fig:PO}
\end{figure}

\paragraph{Centroid-based optimization objective}
\label{sec:objective}
As shown in Figure~\ref{fig:PO}, initial manual homographies often result in 
significant spatial disagreement when the same object is viewed from different 
perspectives. To resolve this, calibration points are grouped into clusters 
$\{\mathcal{C}_c\}$, where each cluster corresponds to one physical landmark 
observed in two or more cameras. Using the original manual projection 
matrices $\homo_i^{(0)}$, we compute an initial BEV centroid for each cluster:
\begin{equation}
  \boldsymbol{\mu}_c^{(0)} = \frac{1}{N_c}\sum_{p\in\mathcal{C}_c}\mathbf{x}_p^{(0)},
\end{equation}
where $N_c = |\mathcal{C}_c|$, $p$ represents an individual image-space observation (e.g., a bounding box foot point from a specific camera) belonging to cluster $\mathcal{C}_c$, and $\mathbf{x}_p^{(0)}$ is its corresponding BEV projection under the original matrices.

\paragraph{Optimization over camera matrices}
Because the system continuously adapts on-the-fly, the projection matrix for each camera is updated iteratively. The optimization cost at iteration $k$ is defined as the sum of the squared distances from each projected point to the reference centroid of its cluster:
\begin{equation}
  \mathcal{L}^{(k)}
    = \sum_c \sum_{p \in \mathcal{C}_c}
      \left\lVert \mathbf{x}_p^{(k)}
                  - \boldsymbol{\mu}_c^{(k_{\mathrm{ref}})}
      \right\rVert^2.
  \label{eq:po_loss}
\end{equation}
An optimizer such as Adam is applied to the current camera matrix parameters $\homo_i^{(k)}$ to minimize this loss $\mathcal{L}^{(k)}$ across all clusters and calibration points. The centroid-based loss naturally extends from two cameras to $N$ cameras: any number of views that contribute points to a cluster are all pulled toward the same cluster centroid in BEV space, yielding the updated matrices $\homo_i^{(k+1)}$ for the next iteration.

\paragraph{Slowly updated centroids}
To avoid trivial collapse and excessive drift, the reference centroids $\boldsymbol{\mu}_c^{(k_{\mathrm{ref}})}$ used in Equation~\ref{eq:po_loss} are refreshed only intermittently. Initially, $\boldsymbol{\mu}_c^{(k_{\mathrm{ref}})}=\boldsymbol{\mu}_c^{(0)}$. Every $K$ optimization iterations we recompute:
\begin{equation}
  \boldsymbol{\mu}_c^{(k_{\mathrm{ref,new}})}
  = \frac{1}{N_c}\sum_{p\in\mathcal{C}_c}\mathbf{x}_p^{(k)}.
  \label{eq:po_centroid_refresh}
\end{equation}

\subsection{Multi-Camera Tracking in BEV}
\label{sec:mv_tracking}

\paragraph{Design principle: a shared global track pool in BEV}
The multi-camera (MC) tracker maintains a \emph{single} set of global tracks whose latent state
lives in a shared bird's-eye-view (BEV) plane. Each camera acts as an independent measurement
source that can update any global track.

\paragraph{Per-camera inputs}
At each timestamp $t$, each camera $i$ produces a set of detections
$\mathcal{D}_t^{(i)}=\{d_{t,m}^{(i)}\}_{m=1}^{M_i}$ with:
(i) a 2D image box and confidence score,
(ii) a chosen image point $\mathbf{p}_{t,m}^{(i)}$ that approximates ground contact (e.g. 
``leg point'' / bottom-center), and
(iii) an appearance embedding $\mathbf{e}_{t,m}^{(i)}$.
The image point is projected to BEV using the per-camera projection model
(Section~\ref{sec:proj_init}--\ref{sec:po}), which yields BEV measurements
$\mathbf{z}_{t,m}^{(i)} \in \mathbb{R}^2$.

\paragraph{Refining Leg Points via Single-View Tracking}
Since our system relies on planar homographies rather than full 3D calibration, multi-camera association is highly sensitive to input noise; a few pixels of vertical jitter in the image plane can translate into large spatial displacements in the BEV. We mitigate this by feeding the multi-camera tracker with \emph{smoothed} state estimates from the single-view tracker rather than raw detections. This temporal stabilization reduces projection variance and provides better-calibrated uncertainty covariances, significantly improving the recall of cross-camera matching under tight Mahalanobis gates.

\subsubsection{Per-camera association against the shared pool}

\paragraph{Parallel matching}
For each timestamp $t$, we first predict all global tracks to obtain
$\{\mathbf{x}_{t|t-1}^{(j)}, \mathbf{P}_{t|t-1}^{(j)}\}_j$.
For each camera $i$, we then independently match its BEV measurements
$\{\mathbf{z}_{t,m}^{(i)}\}_m$ to the \emph{same} global track pool. This step can be
parallelized across cameras, allowing for efficient multi-sensor fusion.

\paragraph{Motion cost (Mahalanobis)}
For detection $m$ at time $t$ in camera $i$ and track $j$, define innovation and covariance:
\begin{equation}
  \mathbf{f}_{m}^{(i,j)} = \mathbf{z}_{t,m}^{(i)} - \mathbf{H}\mathbf{x}_{t|t-1}^{(j)}, \quad
  \mathbf{S}^{(j)} = \mathbf{H}\mathbf{P}_{t|t-1}^{(j)}\mathbf{H}^\top + \mathbf{R}^{(j)},
\end{equation}
where $\mathbf{R}^{(j)}$ may be adapted over time (below). The motion distance is
\begin{equation}
  d_{\text{mot}}(m,j) = (\mathbf{f}_{m}^{(i,j)})^\top (\mathbf{S}^{(j)})^{-1} \mathbf{f}_{m}^{(i,j)}.
  \label{eq:mv_mahal}
\end{equation}

\paragraph{Appearance cost}
We use Euclidean distance between the detection
embedding $\mathbf{e}_{t,m}^{(i)}$ and the track embedding $\mathbf{e}^{(j)}$. 
Embeddings are computed following standard Re-ID practices~\cite{luo2019bag}.

\paragraph{Combined cost and assignment}
Since MC tracks a BEV point (not a 2D box), we do not use IoU-based fallback matching. We construct a combined cost $c(m,j) = \alpha \, d_{\text{mot}}(m,j) + \beta \, d_{\text{app}}(m,j)$. After applying gating thresholds for both motion and total cost, we resolve the one-to-one assignment for camera $i$ using the Hungarian algorithm~\cite{kuhn1955hungarian}.

\subsubsection{Cross-camera fusion into the global state}
After parallel matching, a global track $j$ may receive concurrent measurements from multiple cameras. We fuse these observations via sequential Kalman measurement updates~\cite{brown1997introduction}, where the posterior state and covariance from one camera's update immediately serve as the prior for the next. To handle heterogeneous camera error (e.g., varying projection variance or detector noise), we adapt the measurement noise covariance $\mathbf{R}^{(j)}$ online based on recent residual magnitudes. Finally, the global track's appearance embedding $\mathbf{e}^{(j)}$ is updated via an exponential moving average (EMA) over the matched multi-view detections.

\subsubsection{Track initiation and lifecycle}
\paragraph{Unmatched detections and new track creation}
After association and fusion, each camera may have unmatched detections. We reuse the same
logic as single-view for new track initiation: only detections above a class-specific
\emph{opening threshold} can spawn new tracks. This prevents low-confidence clutter from
creating global IDs. Tracks transition from tentative to confirmed on the basis of configured
hit/miss logic.

\subsection{Trajectory-Level Cross-Camera Track Unification}
\label{sec:mv_collate}

To resolve temporary track fragmentation (e.g., during cross-camera handovers or severe occlusions), we run a trajectory-level merge stage after the per-frame fusion update. We first filter candidates using a strict physical gate: tracks observed by the \emph{same camera} at the \emph{same timestamp} are assigned infinite cost, preventing the collapse of distinct neighboring objects into a single identity.

For admissible candidate pairs $(i,j)$ with a set of overlapping timestamps $\Omega_{ij}$, we aggregate their historical trajectories to assess similarity. Rather than relying on a single frame, we compute the average relative state difference $\bar{\mathbf{d}}_{ij}$ and the average combined covariance $\bar{\boldsymbol{\Sigma}}_{ij}$ across the entire temporal overlap. The structural motion cost is then evaluated via the Mahalanobis distance $d^{\mathrm{motion}}_{ij} = \bar{\mathbf{d}}_{ij}^{\top}\,\bar{\boldsymbol{\Sigma}}_{ij}^{-1}\,\bar{\mathbf{d}}_{ij}$. 

This motion cost is fused with a cosine distance $d^{\mathrm{emb}}_{ij}$ between the tracks' appearance embeddings. Pairs exceeding configured spatial or appearance thresholds are discarded, and the remaining valid pairs are resolved via greedy assignment. Upon merging, the retained track absorbs the duplicate, updating its appearance embedding via a weighted average proportional to the number of multi-view detections each track has accumulated.

\subsection{Implementation Details}
\label{sec:implementation}

\paragraph{Initial Homography Estimation}
The initial projection matrices $\mathbf{A}_i^{(0)}$ are established via a dedicated annotation interface that displays the camera view alongside a 2D site map. Operators select a minimum of four corresponding point pairs (anchors) between the image plane and the bird's-eye-view (BEV) map, deliberately distributing these points as widely as possible across the field of view to maximize coverage and minimize extrapolation errors. The exact number of annotated points is strictly scene-dependent, varying based on the availability of distinct physical landmarks. To ensure a robust initialization, the tool provides a real-time interactive preview: as the operator hovers over the image, the corresponding BEV projection is actively displayed. This feedback loop allows users to iteratively add, move, or delete anchor points until a satisfactory baseline alignment is achieved.

\paragraph{Re-ID Model and Matching Policy}
For cross-camera visual association in the live deployment (Section~\ref{sec:point_acquisition}), the system utilizes a PLR-OSNet \cite{xie2020learning} backbone to extract 2560-dimensional appearance embeddings. Given a frame processing rate of 5~fps, object detections from different cameras are temporally synchronized by matching their metadata timestamps within a narrow tolerance window of $\pm 100$~ms. To prioritize precision over recall during point acquisition, cross-camera associations are evaluated using Euclidean distance and require a strict similarity threshold.

\paragraph{Outlier Rejection}
To further mitigate false Re-ID matches and prevent erroneous associations from corrupting the geometric alignment, we enforce a strict, configurable spatial gating mechanism. Any projected point that yields a spatial distance greater than a defined threshold from its calculated cluster centroid in the shared bird's-eye-view space is rejected as an outlier and excluded from the optimization constraints.

\paragraph{Optimization Parameters and Scheduling}
The projection optimization is performed using the Adam optimizer. Because the BEV projection is highly sensitive to small perturbations in the homography matrix---where minor parameter shifts can cause large spatial displacements and lead to divergence---we employ a very low learning rate of $1 \times 10^{-5}$. We run the optimization for 300,000 epochs; because it optimizes only a minimal parameter space (a $3 \times 3$ matrix per camera) over lightweight metadata, this executes rapidly. To stabilize convergence, the reference cluster centroids (Equation~\ref{eq:po_centroid_refresh}) are updated every 15,000 epochs, but these updates are intentionally frozen after the first 30,000 epochs (10\% of the schedule) to allow the matrix parameters to settle into a stable geometric state. Furthermore, to avoid any computational bottleneck, this optimization service runs asynchronously (e.g., once per day) on a separate machine using the already-transmitted metadata. This architecture ensures that continuous geometric refinement adds zero computational overhead or latency to the real-time multi-view tracking pipeline.

\section{Experiments}
\label{sec:experiments}

\subsection{Setup and Datasets}
We evaluated on three proprietary multi-camera datasets collected in real retail environments. Due to privacy and business constraints, the raw videos and annotations cannot be publicly released. All datasets are recorded at 5 FPS with fixed cameras and are time-synchronized. The ground truth is labeled with global identities across all cameras: the same person observed in multiple views is assigned the same GT track ID.

\paragraph{Dataset A (Supermarket)}
This dataset contains 11 cameras and spans 2:20 minutes. Cameras are installed at approximately 45$^\circ$ view angle and around 3 m height. It includes 53 global ground-truth tracks.

\paragraph{Dataset B (Mall)}
This dataset contains 5 cameras and spans 3:50 minutes. Cameras are installed at approximately 40$^\circ$ view angle and around 4.5 m height. It includes 80 global ground-truth tracks.

\paragraph{Dataset C (General store)}
This dataset contains 8 cameras and spans 2:50 minutes. Cameras are installed at approximately 45$^\circ$ view angle and around 4 m height. It includes 45 global ground-truth tracks.

\subsection{Evaluation Metrics}
We report multi-object tracking metrics (e.g., IDF1 / \allowbreak HOTA / \allowbreak MOTA)~\cite{ristani2016performance,luiten2021hota,bernardin2008evaluating}, ID
switches, and calibration/alignment metrics.
\begin{itemize}
  \item \textbf{BEV alignment error:} for each landmark cluster, compute its BEV centroid as the
mean of all projected points in that cluster; report the mean $\ell_2$ distance from each
projected point to its cluster centroid, averaged over all points across all clusters.
  \item \textbf{Tracking:} IDF1, HOTA; ID switches.
\end{itemize}

\subsection{Baselines and Evaluation Strategy}
Our experimental objective is to isolate and quantify the impact of geometric calibration on multi-camera association. Therefore, rather than comparing disparate tracking architectures—which conflates calibration quality with variations in data association heuristics—we evaluate the \emph{identical} multi-camera tracking pipeline under two different geometric conditions:

\begin{itemize}
  \item \textbf{Manual homographies (no PO).}
  We use the operator-provided per-camera homographies as-is to project detections into BEV and run the multi-camera tracker. This baseline measures how much tracking performance is lost when a standard system relies solely on raw, unrefined manual calibration.

  \item \textbf{Manual + PO (ours).}
  Starting from the exact same manual homographies, we run projection optimization to refine the per-camera projections, and then run the identical multi-camera tracker. By keeping the tracking logic completely static, any improvement over the previous baseline is strictly attributable to better cross-camera BEV alignment.
\end{itemize}

\subsection{Results}

\paragraph{Quantitative Analysis and the Impact of Alignment}
Quantitative results for Datasets A, B, and C are presented in Table~\ref{tab:combined_results}. Across all environments, replacing raw manual homographies with our Projection Optimization (PO) yields consistent improvements in both geometric fidelity and tracking consistency. 

The direct correlation between BEV alignment error and tracking stability is profoundly demonstrated when comparing the dynamics of Datasets~B and~C. Prior to optimization, both datasets suffered from a substantial BEV alignment error of 42~cm. However, the manifestation of this error depends heavily on scene density. Dataset~B represents a busy mall environment with high pedestrian density (80 unique trajectories). In crowded scenes, a 42~cm projection discrepancy is catastrophic: it not only pushes cross-camera observations outside their correct Mahalanobis matching gate (Equation~\ref{eq:mv_mahal}), but frequently pushes them into the spatial gates of neighboring pedestrians. This leads to rampant identity mixing, resulting in an extremely high identity switch count (182 IDSW) under the manual baseline. By applying our centroid-anchored PO, the alignment error in Dataset~B is reduced threefold to 14~cm. This spatial convergence resolves the identity collisions, directly driving the dramatic reduction in ID switches down to 38, while boosting IDF1 from 0.717 to 0.892.

Interestingly, Dataset~C presents a different but equally important insight. Despite starting with the same 42~cm error, its lower target density (a general store with 45 trajectories) means that spatial offsets rarely caused projections to collide with neighboring identities. Consequently, the raw ID switch count remained relatively stable (25 to 24). However, the 42~cm error still caused the tracker to drop associations and break trajectories. By reducing the alignment error from 42~cm to 24~cm, PO ensures that targets maintain their correct global identity for much longer durations without fragmenting, which is reflected in the significant improvements in both IDF1 (0.806 to 0.906) and HOTA (0.683 to 0.810). This confirms that mitigating projection scatter is the primary mechanism for maintaining global trajectory purity, regardless of scene density.

%================================================
\begin{table}[htbp]
\centering
\caption{Quantitative Tracking and Alignment Results.}
\label{tab:combined_results}
\setlength{\tabcolsep}{3pt} % Reduces padding between columns
\renewcommand{\arraystretch}{1.05}
\resizebox{\linewidth}{!}{% Forces table to fit column width
\begin{tabular}{llcccc}
\toprule
Dataset & Method & IDF1 $\uparrow$ & HOTA $\uparrow$ & IDSW $\downarrow$ & AlignErr $\downarrow$ \\
\midrule
\multirow{2}{*}{\shortstack[l]{A}} & Manual & 0.938 & 0.864 & 7 & 37 cm \\
& + PO (ours) & \textbf{0.966} & \textbf{0.926} & \textbf{0} & \textbf{20 cm} \\
\midrule
\multirow{2}{*}{\shortstack[l]{B}} & Manual & 0.717 & 0.544 & 182 & 42 cm \\
& + PO (ours) & \textbf{0.892} & \textbf{0.783} & \textbf{38} & \textbf{14 cm} \\
\midrule
\multirow{2}{*}{\shortstack[l]{C}} & Manual & 0.806 & 0.683 & 25 & 42 cm \\
& + PO (ours) & \textbf{0.906} & \textbf{0.810} & \textbf{24} & \textbf{24 cm} \\
\bottomrule
\end{tabular}%
}
\end{table}

\begin{figure}[htbp]
    \centering
    \begin{minipage}{0.48\linewidth}
        \centering
        \includegraphics[width=\linewidth]{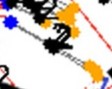}
        \centerline{(a) Before Optimization}
    \end{minipage}\hfill
    \begin{minipage}{0.48\linewidth}
        \centering
        \includegraphics[width=\linewidth]{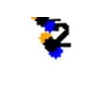}
        \centerline{(b) After Optimization}
    \end{minipage}
    \caption{Detailed anatomy of a single spatial cluster before (a) and after (b) Projection Optimization. In this instance, the blue point represents a target's projection from Camera A, while the yellow point represents the exact same target at the same timestamp from Camera B. The central black point designates the computed cluster centroid, with the  grey lines visualizing the spatial distance from each camera's projection to this shared center. The number ``2'' denotes that this specific cluster fuses observations from two overlapping cameras. Crucially, while this highlights a single synchronized instance, the full-scale maps (Figure~\ref{fig:macro_maps}) plot all clusters generated across the entire video timeline for all people. Because the optimization is agnostic to time and global identity—focusing purely on pulling corresponding multi-view points toward their local centroids—these macroscopic maps effectively visualize the complete geometric input space and convergence objective of the PO algorithm.}
    \label{fig:zoomed_cluster}
\end{figure}

\begin{figure}[htbp]
    \centering
    \begin{minipage}{0.48\linewidth}
        \centering
        \includegraphics[width=\linewidth]{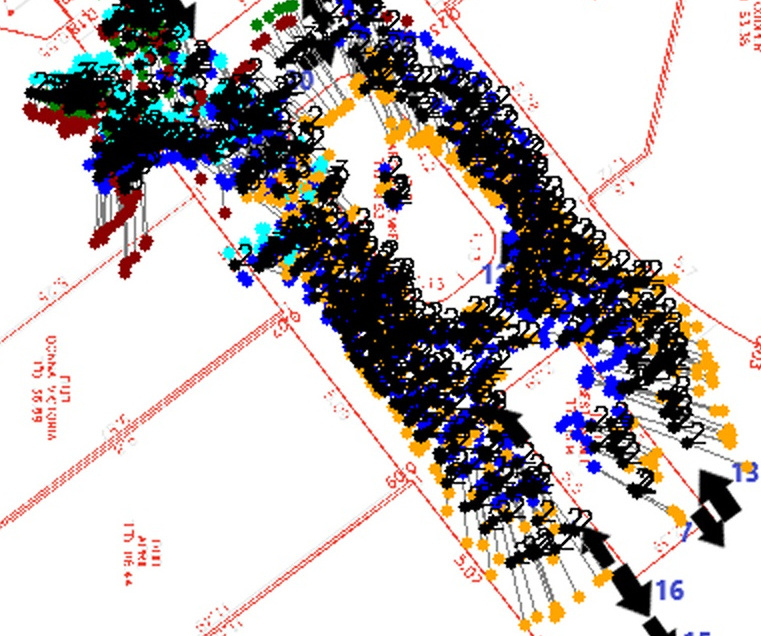}
        \centerline{(a) Before Optimization}
    \end{minipage}\hfill
    \begin{minipage}{0.48\linewidth}
        \centering
        \includegraphics[width=\linewidth]{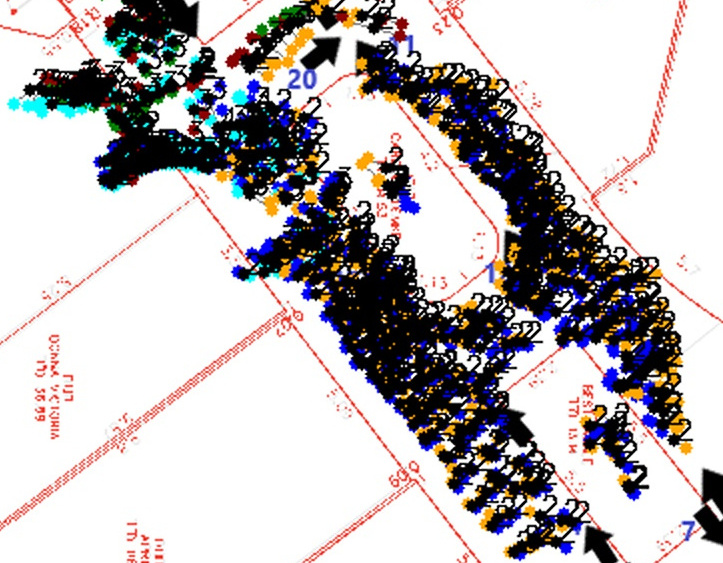}
        \centerline{(b) After Optimization}
    \end{minipage}
    \caption{Global BEV projection consistency across the entire video timeline before (a) and after (b) optimization.}
    \label{fig:macro_maps}
\end{figure}

% ============================================================
\section{Discussion and Limitations}
\label{sec:limits}

\paragraph{Planar Ground Assumption}
Our homography-based formulation strictly assumes that all targets move on a dominant ground plane ($z=0$). Although this approximation holds for the majority of retail and indoor environments, the projection model is inherently limited in scenes with significant elevation changes, such as ramps or split-level flooring. In such non-planar scenarios, a single homography cannot accurately map foot points to the bird's-eye view. Future work could address this by modeling complex sites as piecewise-planar spaces, or by replacing the rigid planar matrix with a lightweight neural projection model capable of learning non-linear, terrain-aware mappings directly from the sparse detection metadata.

\paragraph{Dependence on View Overlap}
The efficacy of Projection Optimization (PO) relies on the existence of a connected graph of overlapping views. The optimization constraints are derived entirely from joint observations—either shared static landmarks or clustered detections visible in multiple cameras. If a camera is spatially isolated (i.e., it shares no common field of view with the network), its projection parameters cannot be geometrically refined relative to the global frame. In these disjoint cases, the system safely falls back to the initial manual calibration, relying more on the visual Re-ID embeddings rather than spatial gating for cross-zone associations.

\paragraph{Initialization Sensitivity}
Finally, centroid-based optimization is designed as a local refinement step rather than a global geometric solver. It assumes that the initial manual homographies provide a roughly correct starting point regarding scale and orientation. If the initial manual input is grossly incorrect (e.g., flipped axes or massive scale errors), the optimization may converge to a degenerate local minimum. However, in practice, the real-time visual feedback provided by our annotation UI (Section~\ref{sec:implementation}) effectively prevents these severe initialization errors, ensuring the starting geometry is well within the convergence basin of the PO algorithm.

% ============================================================
% \FloatBarrier
\section{Conclusion}
\label{sec:conclusion}
We presented a calibration-light multi-camera tracking framework that successfully bridges the gap between geometric precision and real-world deployment constraints. By introducing a centroid-anchored Projection Optimization (PO) method, we demonstrated that coarse, manual planar homographies can be dynamically refined into a robust spatial coordinate system using exclusively live, anonymous detection metadata. This continuous, on-the-fly alignment eliminates multi-view projection scatter, restoring the integrity of spatial distance gating and significantly reducing global identity switches. 

Furthermore, because the optimization operates asynchronously on lightweight metadata, it completely decouples geometric self-healing from the real-time tracking loop, incurring zero computational overhead. Ultimately, our approach yields a highly scalable, privacy-compliant pipeline capable of maintaining stable global trajectories in dynamic environments where traditional static 3D calibration and centralized video processing are fundamentally prohibitive or infeasible.

% ============================================================
% ---------- References ----------

% Use your conference bib style here.
\bibliographystyle{plain}
\bibliography{references}

\end{document}